\def\year{2022}\relax
\documentclass[letterpaper]{article} 
\usepackage{aaai22}  
\usepackage{times}  
\usepackage{helvet}  
\usepackage{courier}  
\usepackage[hyphens]{url}  
\usepackage{graphicx} 
\usepackage{natbib}  
\usepackage{caption} 
\DeclareCaptionStyle{ruled}{labelfont=normalfont,labelsep=colon,strut=off} 
\usepackage{algorithm}
\usepackage{algorithmic}

\usepackage{newfloat}
\usepackage{listings}
\floatstyle{ruled}
\newfloat{listing}{tb}{lst}{}
\floatname{listing}{Listing}
\usepackage{amssymb,amsmath,amsthm,latexsym}
\renewcommand{\citename}{\citet}
\renewcommand{\cite}{\citep}
\graphicspath{{./figures/}}

\newcommand{\eg}{e.g.,~}
\newcommand{\ie}{i.e.,~}
\usepackage{sg-macros}

\title{Exploiting Problem Structure in Deep Declarative Networks: Two Case Studies}

\author {
Stephen Gould\textsuperscript{\rm 1},
Dylan Campbell\textsuperscript{\rm 2},
Itzik Ben-Shabat\textsuperscript{\rm 1,3},\\
Chamin Hewa Koneputugodage\textsuperscript{\rm 1},
Zhiwei Xu\textsuperscript{\rm 1}
}
\affiliations {
\textsuperscript{\rm 1} Australian National University, Canberra, Australia\\
\textsuperscript{\rm 2} University of Oxford, Oxford, United Kingdom\\
\textsuperscript{\rm 3} Technion, Haifa, Israel
}

\begin{document}

\maketitle

\begin{abstract}
	Deep declarative networks and other recent related works have shown how to
	differentiate the solution map of a (continuous) parametrized optimization problem,
	opening up the possibility of embedding mathematical optimization problems into
	end-to-end learnable models. These differentiability results can lead to significant memory savings by providing an expression for computing the derivative without needing to unroll the steps of the forward-pass optimization procedure during the backward pass.
	However, the results typically require inverting a large
	Hessian matrix, which is computationally expensive when implemented naively.
	In this work we study two applications of deep declarative networks---robust vector pooling and optimal transport---and
	show how problem structure can be exploited to obtain very efficient backward pass
	computations in terms of both time and memory. Our ideas can be used as a guide for improving the computational performance of other novel deep declarative nodes.
\end{abstract}


\section{Introduction}
\label{sec:intro}

Deep declarative networks, also known as differentiable optimization or implicit layers~\cite{Gould:PAMI2021, Agrawal:NIPS2019, Amos:ICML2017}, are deep learning models that support propagating (exact) gradients backwards through the solution of a continuous optimization problem. This is achieved by applying the implicit function theorem to the optimality conditions of the problem at a given solution. The advantage of this approach is that intermediate results produced by the (typically iterative) optimization algorithm need not be cached for use in the backward pass. Indeed, non-differentiable steps can be applied during the forward pass and details of the optimization algorithm do not even need to be known for calculating the gradient in the backward pass.

Specifically, an expression for the Jacobian $\dd y(x)$ of the output $y$ with respect to the input $x$ can be formulated knowing only the optimality conditions for the problem at hand and the current solution. Moreover, given a software implementation of the objective and constraints (or the optimality condition directly) the gradient can be computed without additional coding by automatic differentiation~\cite{PyTorch, Blondel:TR2021}. However, notwithstanding the significant savings in development time, automatic differentiation can in some situations lead to suboptimal computations, and implemented poorly the result may be even slower and more memory intensive than unrolling and back-propagating through the forward pass optimization loop.

The core operation performed by a deep learning node or layer during the backward pass is to calculate the gradient of the loss function (or global objective) $\dd J(x)$ with respect to the node's inputs (or parameters) given the gradient of the loss function with respect to its outputs $\dd J(y)$. The calculation is an instance of the chain rule for differentiation:
\begin{align}
	\dd J(x) &= \dd J(y) \cdot \dd y(x),
	\label{eqn:chain_rule}
\end{align}
where $J$ is the loss function and $\dd y(x)$ is the gradient of the output with respect to the input. In PyTorch this is the role of the \texttt{backward} method of \texttt{autograd.Function}~\cite{PyTorch} that then allows gradients to back-propagate through the entire network.

\citename{Gould:PAMI2021} consider deep declarative nodes defined by second-order differentiable, equality constrained, optimization problems parametrized by an $n$-dimensional input $x$ of the form
\begin{align}
	\arraycolsep=2pt
	\begin{array}{rllr}
		y(x) \in& {\textstyle\argmin_{u \in \reals^m}} & f(x, u) &\\
		& \text{subject to} & h_i(x, u) = 0, & i = 1, \ldots, p
	\end{array}
\end{align}
and give an expression for $\dd y(x)$ as
\begin{align}
	H^{-1}A\transpose\left(AH^{-1}A\transpose\right)^{-1}\left(AH^{-1}B - C\right) - H^{-1}B,
	\label{eqn:dydx}
\end{align}
where $A$, $B$, $C$ and $H$ are objects (matrices or tensors) of first- and second-order (mixed) partial derivatives of the objective and constraint functions with respect to $x \in \reals^n$ and $y \in \reals^m$.
Specifically,
\begin{align*}
	A &= \dd[Y] h(x,y) \in \reals^{p \times m}\\
	B &= \dd[XY]^2 f(x, y) - \sum_{i = 1}^{p} \lambda_i \dd[XY]^2 h_i(x, y) \in \reals^{m \times n}\\
	C &= \dd[X] h(x,y) \in \reals^{p \times n}\\
	H &= \dd[YY]^2 f(x, y) - \sum_{i = 1}^{p} \lambda_i \dd[YY]^2 h_i(x, y)  \in \reals^{m \times m}
\end{align*}
and $\lambda \in \reals^p$ satisfies $\lambda\transpose \!A = \dd[Y] f(x, y)$.
Here, the notation comes from \citename{Gould:PAMI2021} with $\dd[Z]$ denoting partial derivatives with respect to variables $Z$. Naive implementation of \eqnref{eqn:dydx} requires $O(\max\{m^3, p^3\})$ operations due to the matrix inversions.

In general, the loss function $J$ is scalar-valued and summed over each training example in a mini-batch. As such gradients of $J$ with respect to each node's inputs and outputs decompose over elements of the mini-batch, and Equations~\ref{eqn:chain_rule}~and~\ref{eqn:dydx} can be evaluated independently (and in parallel) for each training example of the mini-batch. Let $b$ be the size of the mini-batch, $n$ be the size of the input and $m$ be the size of the output. Then storage for $\dd J(y)$,  $\dd y(x)$, and $\dd J(x)$ requires $O(bm)$, $O(bnm)$ and $O(bn)$ bytes, respectively. However, for many optimization problems we do not need to construct $\dd y(x)$ explicitly and can instead exploit its structure to save both computation and memory.

Deep declarative networks provide a powerful and flexible tool that has been applied to a growing number of applications including video classification~\cite{Fernando:CVPR2016}, visual Sudoku~\cite{Amos:ICML2017, Wang:ICML2019}, blind PnP~\cite{Campbell:ECCV2020, Chen:CVPR2020} and meta-learning~\cite{Lee:CVPR2019}. The contribution of this paper is to provide case studies that demonstrate general principles for implementing efficient backward pass computation in deep declarative nodes so as not to be a bottleneck. Based on the case studies and our experience, we conclude with tips and advice for implementing new declarative nodes.


\section{Background and Related Work}
\label{sec:background}

Automatic differentiation is the backbone of modern deep learning software frameworks such as PyTorch~\cite{PyTorch}. It allows rapid experimentation with different network architectures and implementation of new differentiable processing nodes, where the forward pass can be explicitly implemented as a sequence of steps, themselves differentiable expressions. Deep declarative networks~\cite{Gould:PAMI2021} introduced a new form of processing node as the solution to an optimization problem, where the algorithm for implementing the forward pass is not explicitly defined, but where back-propagation through the node is still possible.

Early examples of such declarative nodes in deep networks~\cite{Amos:ICML2017, Gould:TR2016, Fernando:CVPR2016} relied on hand-coded implementations of the backward pass. Later works show that automatic differentiation techniques can also be applied in the case of deep declarative nodes by differentiating the optimality conditions for the problem at hand~\cite{Agrawal:NIPS2019, cvxpy, Gould:PAMI2021, Blondel:TR2021}, dramatically simplifying the implementation of these nodes. However, this automatic approach is less able to exploit structure that may exist in the problem, and as a result is suboptimal. Thus, it is sometimes desirable to revert to carefully crafted manual implementations.

Early work that exploits problem structure includes \citename{Fernando:ICML2016} for the case of differentiable rank pooling, where the Sherman--Morrison formula~\cite{Horn:1991} was used to efficiently compute the inverse of a Hessian matrix required during the backward pass. The same work and others suggest applying approximations to simplify the backward pass, \eg taking the diagonal of the Hessian~\cite{Fernando:ICML2016}, ignoring constraints, or heavily regularising to reduce the number of iterations in the forward pass~\cite{Asano:ICLR2021}. We provide further examples showing general patterns for exploiting structure and opportunities for approximation.


\section{Case Studies}
\label{sec:cases}

We present two case studies of deep declarative nodes---one unconstrained and one constrained. The case studies follow a generic recipe for implementing deep declarative nodes: (i)~Write out the mathematical expressions for the objective and constraints; (ii)~Derive the relevant partial derivatives needed in \eqnref{eqn:dydx}; (iii)~Inspect the components for structure and consider how to implement them efficiently; (iv)~Code and test the forward and backward passes. Experiments profiling memory and running time are included for each example, and full PyTorch source code is available.\footnote{All results are reported using PyTorch 1.8.1 with robust vector pooling running on NVIDIA GeForce RTX 2080 GPU and optimal transport on NVIDIA GeForce RTX 3090.}

\subsection{Robust Vector Pooling}
\label{sec:rvp}

Consider the problem of computing a robust estimate for the mean of a set of $m$-dimensional points $\X = \{x_i \in \reals^m \mid i = 1, \ldots, n\}$. That is, we assume that our data $\X$ is noisy and wish to find the point $y \in \reals^m$ that best approximates the mean of the noise-free data. If we knew the noise model then this amounts to solving a maximum-likelihood problem. For example, under an isotropic Gaussian noise model (or no noise) the best approximation is the \emph{sample mean}, $y = \frac{1}{n} \sum_{i=1}^{n} x_i$. In other situations, we may want to reduce the effect of outliers, and do so by finding a point $y$ that minimizes the sum of costs for the distance to each point $x_i$,
\begin{align}
	y \in \text{arg min}_{u \in \reals^m} \sum_{i=1}^{n} \phi(\|u - x_i\|_2; \alpha),
	\label{eqn:rvp}
\end{align}
where $\phi: \reals \to \reals_{+}$ is a penalty function parametrized by $\alpha$. For the one-dimensional case ($m=1$) this is an instance of the \emph{penalty function approximation problem}~\cite{Boyd:2004}. When using a quadratic penalty function, $z \mapsto \frac{1}{2} z^2$, the solution is the sample mean. However, this is not robust to outliers and many other penalty functions have been proposed (\eg see \tabref{tab:penalties}).%
\footnote{Note that \citename{Gould:PAMI2021} consider the one-dimensional case, applying the penalty function to $u - x_i$, which is computationally more straightforward since $H$ and $B$ are scalars. Here we generalize to the vector case and apply the penalty function to $\|u - x_i\|_2$, which requires more care in implementing operations on $m$-by-$m$ matrices.}

The objective function for the robust vector pooling optimization problem (\eqnref{eqn:rvp}) is
\begin{align}
	f(\X, u) &= \sum_{i=1}^{n} \phi(\|u - x_i\|_2; \alpha)
	= \sum_{i=1}^{n} \phi(z_i; \alpha),
\end{align}
where we have written $z_i = \|u - x_i\|_2$. Since the problem is unconstrained, the gradient of the minimizer $y$ with respect to each of the $x_j$ reduces to~\cite[Proposition 4.4]{Gould:PAMI2021}
\begin{align}
	\dd[X_j] y = -H^{-1} B,
\end{align}
where $H = \dd[YY]^2 f$ and $B = \dd[X_jY]^2 f$. Since $f$ decomposes as a sum of penalty functions $\phi$, it suffices to just consider $\dd[YY]^2 \phi$ and $\dd[XY]^2 \phi$. Let us start by computing $\dd[Y] \phi$ for the $i$-th data point,
\begin{align}
	\dd[Y] \phi(z_i; \alpha) &= \phi'(z_i; \alpha) \dd[Y] z_i
	= \frac{\phi'(z_i; \alpha)}{z_i} (y - x_i)\transpose,
	\label{eqn:rvp_dy}
\end{align}
where $\phi'$ is the first derivative of $\phi$. Computing second derivatives, we have
\begin{align}
	\dd[YY]^2 \phi(z_i; \alpha)
	&= \frac{\phi'(z_i; \alpha)}{z_i} I_{m \times m} + {} \\
	&\quad \left(\frac{\phi''(z_i; \alpha)}{z_i^2} - \frac{\phi'(z_i; \alpha)}{z_i^3}\right) (y - x_i)(y - x_i)\transpose \notag
	\\
	&= \kappa_1(z_i) I_{m \times m} + \kappa_2(z_i) (y - x_i)(y - x_i)\transpose \notag,
\end{align}
where $\kappa_1$ and $\kappa_2$ are quantities that depend on the penalty function and $z_i$ (see \tabref{tab:penalties}). By anti-symmetry of $x_i$ and $y$ in \eqnref{eqn:rvp_dy}, we have $\dd[X_jY]^2 \phi(z_j; \alpha) = - \dd[YY]^2 \phi(z_j; \alpha)$.
We can therefore write the following expression for $\dd[X_j] y$,
\begin{multline}
	\underbrace{\left( \sum_{i=1}^{n} \kappa_1(z_i) I + \kappa_2(z_i) (y - x_i)(y - x_i)\transpose \right)^{-1}}_{H^{-1}} \\
	\underbrace{\Bigg( \kappa_1(z_j) I + \kappa_2(z_j) (y - x_j)(y - x_j)\transpose\Bigg)}_{-B}.
\end{multline}

\begin{table*}[t!]
	\centering
	\footnotesize
	\def\arraystretch{1.5}
	\begin{tabular}{l|ccc}
		& $\phi(z; \alpha)$ & $\kappa_1(z; \alpha)$ & $\kappa_2(z; \alpha)$
		\\
		\hline
		{\sc Quadratic} & $\frac{1}{2} z^2$ & 1 & 0
		\\
		{\sc Pseudo-Huber}
		& $\alpha^2 \left( \sqrt{1 + \left( \frac{z}{\alpha}\right)^2 } - 1 \right)$ 
		& $\left(1 + \left( \frac{z}{\alpha}\right)^2\right)^{-1/2}$
		& $-\frac{1}{\alpha^2} \left(1 + \left( \frac{z}{\alpha}\right)^2 \right)^{-3/2}$
		\\
		{\sc Huber}
		& \(
		\begin{cases}
			\frac{1}{2} z^2 & \text{for $|z| \leq \alpha$} \\
			\alpha(|z| - \frac{1}{2}\alpha) & \text{otherwise}
		\end{cases}
		\)
		& \(
		\begin{cases}
			1 & \text{for $|z| \leq \alpha$} \\
			\alpha / |z| & \text{otherwise}
		\end{cases}
		\)
		& \(
		\begin{cases}
			0 & \text{for $|z| \leq \alpha$} \\
			-\alpha / |z|^3 & \text{otherwise}
		\end{cases}
		\)		
		\\
		{\sc Welsch}
		& $1 - \exp\left(-\frac{z^2}{2 \alpha^2} \right)$
		& $\frac{1}{\alpha^2} \exp\left(-\frac{z^2}{2 \alpha^2} \right)$
		& $-\frac{1}{\alpha^4} \exp\left(-\frac{z^2}{2 \alpha^2} \right)$
		\\
		{\sc Trunc. Quad.}
		& \(
		\begin{cases}
			\frac{1}{2} z^2 & \text{for $|z| \leq \alpha$} \\
			\frac{1}{2} \alpha^2 & \text{otherwise}
		\end{cases}
		\)
		& \(
		\begin{cases}
			1 & \text{for $|z| \leq \alpha$} \\
			0 & \text{otherwise}
		\end{cases}
		\)
		& 0
	\end{tabular}
	\caption{Parameters $\kappa_1$ and $\kappa_2$ for various robust penalty functions $\phi$
		where $\kappa_1(z) = \phi'(z) / z$ and $\kappa_2(z) = (\phi''(z) - \kappa_1(z)) / z^2$.
		In the case of robust vector pooling the argument $z$ is non-negative and the absolute
		value calculations can be omitted.}
	\label{tab:penalties}
\end{table*}

A naive implementation of this expression would be prohibitively expensive since $B$ is an $m$-by-$m$ matrix that must to be computed separately for each point $x_j \in \X$ (or stored if computed in batch during the construction of $H$ requiring $O(nm^2)$ memory).
It is preferable to compute $\dd[X_j] y$ for all $j$ at the same time, \ie in batch, to make use of GPU parallelization, which further exacerbates the memory problem.
A better approach is to evaluate the entire expression for the gradient of the loss function (\eqnref{eqn:chain_rule}) from left-to-right.

Let $v\transpose = \dd J(y)$ be the derivative of the loss function with respect to the output, i.e., the incoming backward gradient. Our goal is to compute $\dd J(x_i)$ for $i = 1, \ldots, n$. We have,
$\dd J(x_i) = v\transpose H^{-1} B$.
Letting $w\transpose = v\transpose H^{-1}$ be obtained by solving $v = H w$ using Cholesky factorization and back substitution. Note that this can be computed once for all points in the input as it is independent of which $x_i$ we are taking the derivative with respect to. We then have
\begin{align}
	\dd J(x_i)
	&= \kappa_1(z_i) w\transpose + \kappa_2(z_i) w\transpose (y - x_i)(y - x_i)\transpose.
	\label{eqn:rvp_result}
\end{align}

Taking the inner product $w\transpose (y - x_i)$ first, instead of the outer product $(y - x_i)(y - x_i)\transpose$, results in significant memory and computational savings, requiring only $O(nm)$ bytes of storage when processed in batch versus $O(nm^2)$. Note also that some penalty functions have $\kappa_2 \equiv 0$, \eg quadratic, thus avoiding this computation entirely (see \tabref{tab:penalties}).

\begin{figure}
	\small
	\setlength{\tabcolsep}{1pt}
	\begin{tabular}{cccc}
		\raisebox{-0.5\height}{\includegraphics[width=0.23\textwidth]{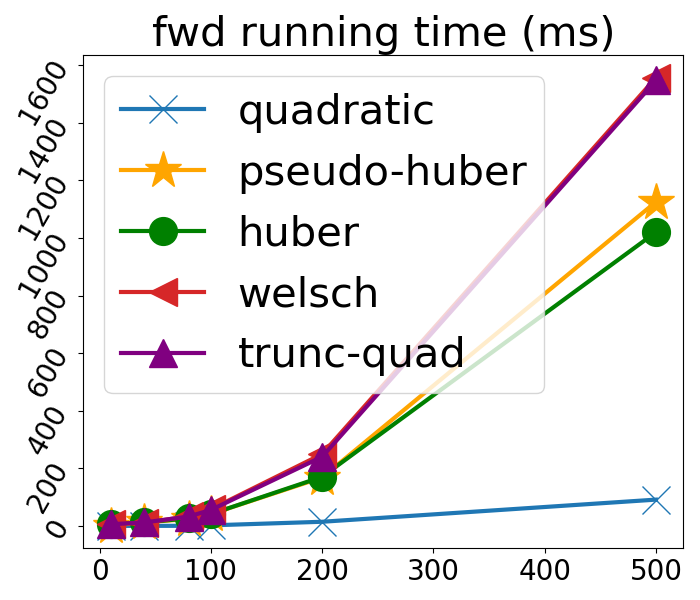}}
		& \raisebox{-0.5\height}{\includegraphics[width=0.23\textwidth]{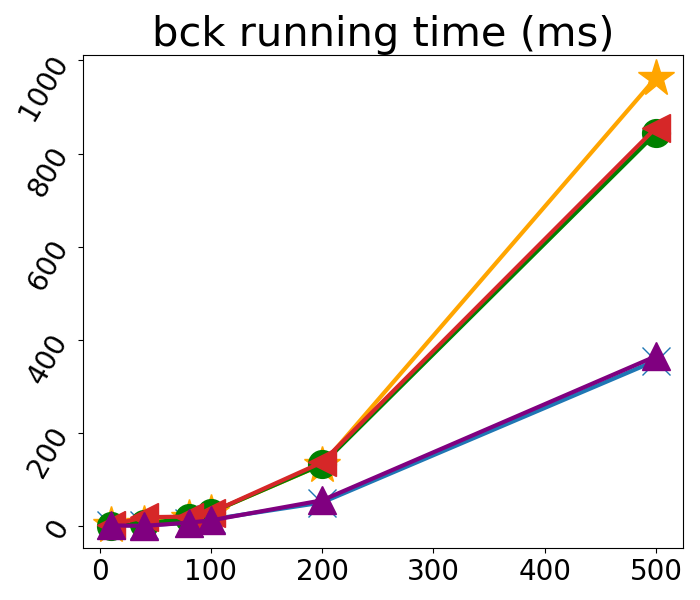}}
		\\
		\raisebox{-0.5\height}{\includegraphics[width=0.23\textwidth]{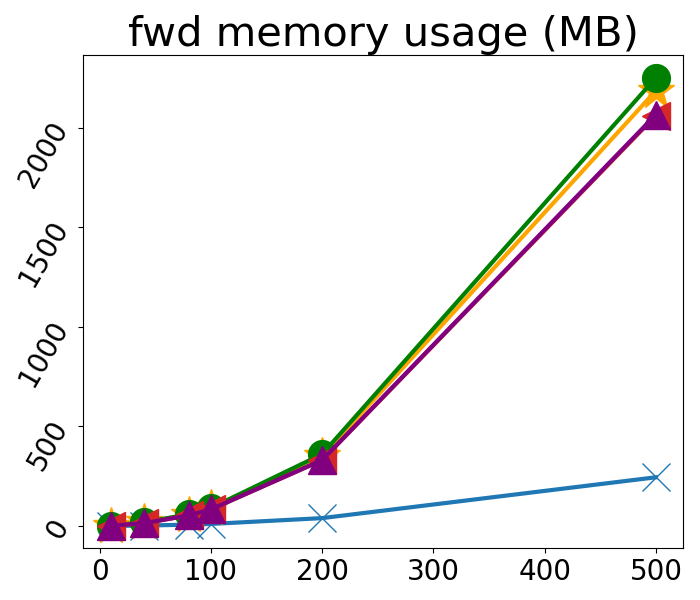}}
		& \raisebox{-0.5\height}{\includegraphics[width=0.23\textwidth]{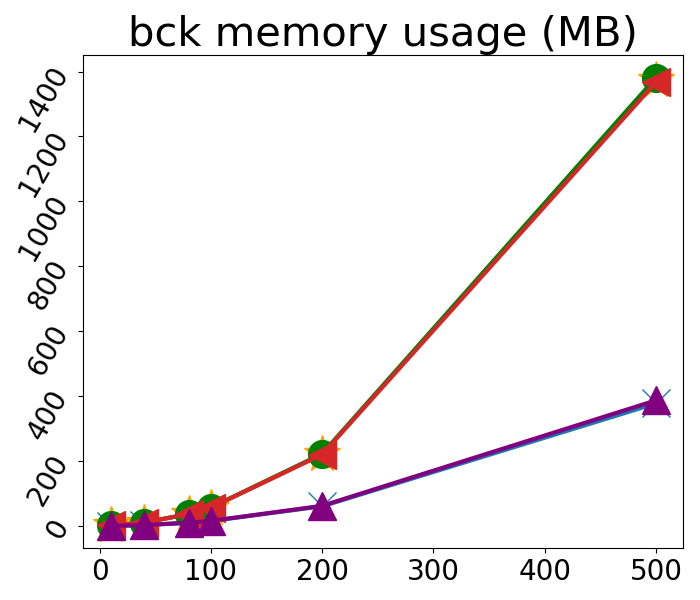}}
	\end{tabular}
	\caption{Time (top) and memory (bottom) requirements for forward and backward passes of robust vector pooling on the CPU.
	The 2D feature map ($\sqrt{n} \times \sqrt{n}$) has $m=128$ channels, and the batch size one.
	We use L-BFGS in the forward pass except for quadratic, which has a closed-form solution. For non-convex penalties we take the best solution from two different initializations. The backward pass is implemented by implicit differentiation following the expression in \eqnref{eqn:rvp_result} (code in \figref{code:robust_pooling}).}
    \label{fig:rvp_runtime_memory}
\end{figure}

\figref{code:robust_pooling} shows PyTorch source code for the backward pass. The code handles both the case of $\kappa_2 = 0$ (Lines 9--10) and the case of $\kappa_2 \neq 0$ (Lines 12--20), and follows a batch implementation of the expression above. Profiling of forward and backward passes for different size problems and different penalty functions is shown in \figref{fig:rvp_runtime_memory}.  Observe that memory for the forward and backward passes is comparable.

\begin{figure*}
\begin{pycode}
def backward(ctx, v):
	x, y = ctx.saved_tensors
	b, m = x.shape[:2]
	
	y_minus_x = y.view(b,m,1) - x.view(b,m,-1)
	z = linalg.norm(y_minus_x, dim=1, keepdim=True) + 1.0e-9
	
	k1, k2 = ctx.penalty.kappa(z, ctx.alpha)
	if all(k2 == 0.0):
		return (k1 * (y_grad / k1.sum(dim=2)).view(b,m,1)).reshape(x.shape)

	H = k1.sum(dim=2).view(b,1,1) * eye(m).view(1,m,m) + \
		einsum("bik,bjk->bij", y_minus_x, k2 * y_minus_x)
	
	L = cholesky(H)
	w = cholesky_solve(v.view(b,m,-1), L).view(b,m)
	
	u = einsum("bi,bik->bk", w, k2 * y_minus_x)
	
	return (k1 * w.view(b,m,1) + einsum("bk,bik->bik", u, y_minus_x)).reshape(x.shape)
\end{pycode}
\caption{Implementation of the backward pass for robust vector pooling. It is assumed that the $b$-by-$m$-by-$n$ input and $b$-by-$m$ output are cached in the forward pass. Global pooling is done over $m$-dimensional features for each of the $b$ batches independently. The function \texttt{ctx.penalty.kappa} computes $\kappa_1$ and $\kappa_2$ for the given penalty function (see \tabref{tab:penalties}). Full source code available at \url{http://deepdeclarativenetworks.com}.}
\label{code:robust_pooling}
\end{figure*}

\subsection{Optimal Transport}
\label{sec:op}

Optimal transport is a very popular algorithm in machine learning for measuring the distance between two probability distributions. It can also be used to find matches between sets of objects (\eg in solving the blind PnP problem~\cite{Campbell:ECCV2020}).
The entropy regularized optimal transport problem can be written as the linearly constrained mathematical program,
\begin{align}
  \begin{array}{ll}
    \text{minimize (over $P \in \reals^{m \times n}_{+}$)} & \langle P, M\rangle + \frac{1}{\gamma} \text{KL}(P \| rc\transpose) \\
    \text{subject to} & P 1 = r \\ & P\transpose 1 = c,
  \end{array}
  \label{eqn:op}
\end{align}
where $M \in \reals^{m \times n}$ is the input cost matrix, $r$ and $c$ are positive vectors of row and column sums (with $\ones\transpose r = \ones\transpose c = 1$), and $\gamma > 0$ controls the strength of the regularization term.
What makes this formulation attractive from a computational perspective is that it can be solved very efficiently by the Sinkhorn algorithm, an iterative algorithm that performs successive row and column normalizations~\cite{Cuturi:NIPS2013}.

In computing derivatives we arrive at similar expressions to \citename{Luise:NIPS2018}, who present an algorithm for differentiating with respect to $r$ and $c$, but where we directly use the results for deep declarative nodes (\eqnref{eqn:dydx}).
To find $\dd[M] P$, the Jacobian of $P$ with respect to $M$, let $f(M, P) = \sum_{ij} M_{ij} P_{ij} + \frac{1}{\gamma} \sum_{ij} P_{ij} \left(\log P_{ij} - \log r_i c_j\right)$ be the entropy regularized optimal transport objective. We can then write the following partial derivatives,
\begin{align}
  \frac{\partial f}{\partial P_{ij}} &= M_{ij} + \frac{1}{\gamma} \log P_{ij} + \frac{1}{\gamma} - \frac{1}{\gamma} \log r_i c_j,
  \\
  H_{ij,kl} &= \frac{\partial^2 f}{\partial P_{ij} \partial P_{kl}} = \begin{cases}
    \frac{1}{\gamma P_{ij}} & \text{if $(i,j) = (k,l)$} \\
    0 & \text{otherwise,}
  \end{cases}
  \\
  B_{ij,kl} &= \frac{\partial^2 f}{\partial P_{ij} \partial M_{kl}} = \begin{cases}
    1 & \text{if $(i,j) = (k,l)$} \\
    0 & \text{otherwise,}
  \end{cases}
\end{align}
for $i = 1, \ldots, m$ and $j = 1, \ldots, n$. Flattening the input $M$ and output $P$ into vectors rowwise we have $H^{-1} = \diag{\gamma P_{ij}}$ and $B = I_{mn \times mn}$. That these are diagonal makes sense, since if not for the linear equality constraints each $P_{ij}$ would only depend on $M_{ij}$ and not any other $M_{kl}$ for $kl \neq ij$. Moreover, since $B$ is the identity matrix we can ignore it from any calculations.

The primary challenge now is computing the term $(AH^{-1}A\transpose)^{-1}$ in \eqnref{eqn:dydx}. Here the matrix $A$ of partial derivatives of the constraint functions with respect to the output is formed as the coefficients of the $P_{ij}$ in the constraint functions of Problem~\ref{eqn:op},
\begin{align}
  h: \left\{
  \begin{array}{ll}
    \sum_{j=1}^{n} P_{ij} - r_i & \text{for } i = 1, \ldots, m \\
    \sum_{i=1}^{m} P_{ij} - c_j & \text{for } j = 1, \ldots, n
  \end{array}
  \right\} = 0.
  \label{eqn:op_constraints}
\end{align}

Note that the set contains a redundant constraint; if any $m + n - 1$ constraints are satisfied then the remaining constraint will also be satisfied. To apply \eqnref{eqn:dydx} we must remove one constraint otherwise $A$ will not be full rank~\cite[Corollary 4.9]{Gould:PAMI2021}. Removing the first constraint and where $P_{ij}$ has again been flattened rowwise, we have
\begin{align}
  A &= \begin{bmatrix}
    0_n\transpose & 1_n\transpose & \cdots & 0_n\transpose \\
    \vdots & \vdots & \ddots & \vdots \\
    0_n\transpose & 0_n\transpose & \cdots & 1_n\transpose \\
    I_{n \times n} & I_{n \times n} & \cdots & I_{n \times n}
\end{bmatrix} \in \reals^{(m+n-1)\times mn}.
\end{align}

It is straightforward to show that
\begin{align}
	AH^{-1}A\transpose
	&=
	\left[
		\begin{matrix}
			\diag{\sum_{j=1}^{n} H_{pj,pj}^{-1} \mid p = 2, \ldots, m} \\
				(H_{ij,ij}^{-1})_{j=1,\ldots,n \times i=2,\ldots,m}
		\end{matrix}
	\right. \notag \\ &\qquad \left.
		\begin{matrix}
			(H_{ij,ij}^{-1})_{i=2,\ldots,m \times j=1,\ldots,n} \\
			\diag{\sum_{i=1}^{m} H_{ip,ip}^{-1} \mid p = 1, \ldots, n}			
		\end{matrix}
	\right]
	\\
	&=
	\gamma \begin{bmatrix}
		\diag{r_{2:m}}
		&
		P_{2:m,1:n}
		\\
		P_{2:m,1:n}\transpose
		&
		\diag{c}
	\end{bmatrix}
\end{align}
by considering the $(p,q)$-th entry of $AH^{-1}A\transpose$ for $p, q \in 1, \ldots, m+n-1$ as,
\begin{align}
  (A H^{-1} A\transpose)_{pq} &= \sum_{i=1}^{m} \sum_{j=1}^{n} \frac{A_{p,ij} A_{q,ij}}{H_{ij,ij}}
\end{align}
and substituting $\gamma P_{ij}$ for $H_{ij,ij}^{-1}$, and $r_{2:m}$ and $c$ for their corresponding sums.

Now we can directly compute $(AH^{-1}A\transpose)^{-1}$ in $O((m+n-1)^3)$ time or make use of more efficient block matrix inversion~\cite{Horn:1991} results to compute in $O((m-1)^3)$ time,\footnote{Or in $O(n^3)$ time if $n < m$ using an alternative formula for the block inverse.}
\begin{align}
\begin{bmatrix}
  \Lambda_{11} & \Lambda_{12} \\ \Lambda_{12}\transpose & \Lambda_{22}
\end{bmatrix}
&=
\begin{bmatrix}
  \diag{r_{2:m}}
  &
  P_{2:m,1:n}
  \\
  P_{2:m,1:n}\transpose
  &
  \diag{c}
\end{bmatrix}^{-1},
\end{align}
where each block is calculated as
\begin{align}
  \Lambda_{11} &= \left(\diag{r_{2:m}} - P_{2:m,1:n} \diag{c}^{-1} \!\! P_{2:m,1:n}\transpose\!\right)^{\!-1} \\
  \Lambda_{12} &= -\Lambda_{11} P_{2:m,1:n} \diag{c}^{-1} \\
  \Lambda_{22} &= \diag{c}^{-1} \left(I - P_{2:m,1:n}\transpose \Lambda_{12}\right)
\end{align}
and we use Cholesky factorization to multiply by $\Lambda_{11}$ rather than inverting explicitly.

\begin{figure}
	\small
	\setlength{\tabcolsep}{1pt}
	\begin{tabular}{cc}
		\includegraphics[width=0.23\textwidth]{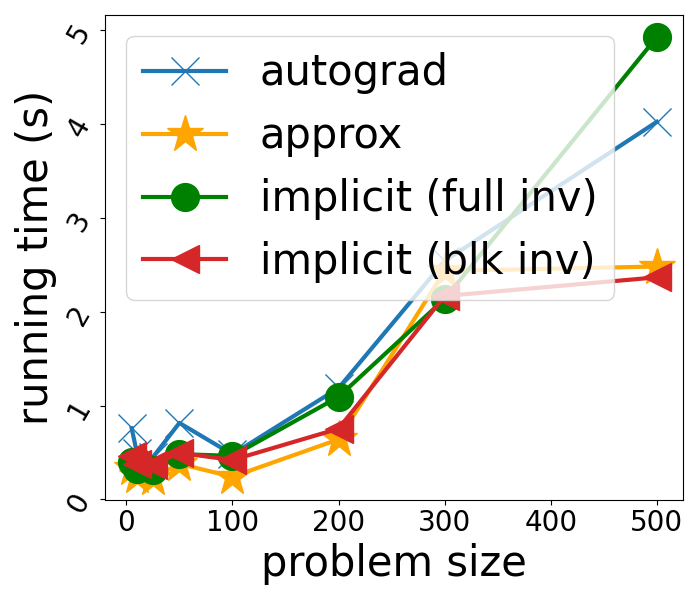}
		& \includegraphics[width=0.23\textwidth]{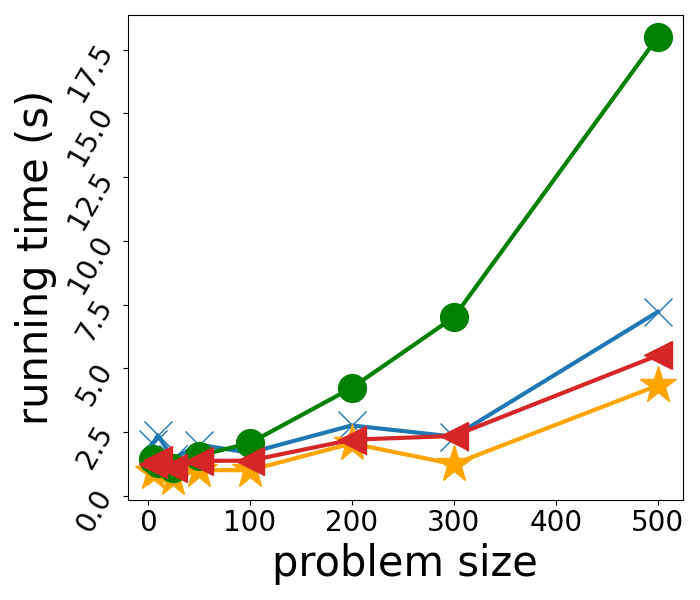}
		\\ (a) CPU {\scriptsize (batch size 1)}
		& (b) GPU {\scriptsize (batch size 16)}
		\\ \includegraphics[width=0.23\textwidth]{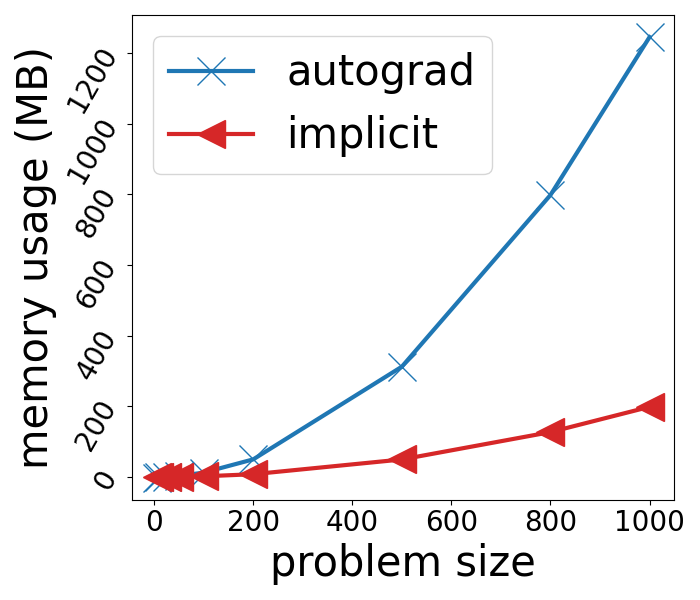}
		& \includegraphics[width=0.23\textwidth]{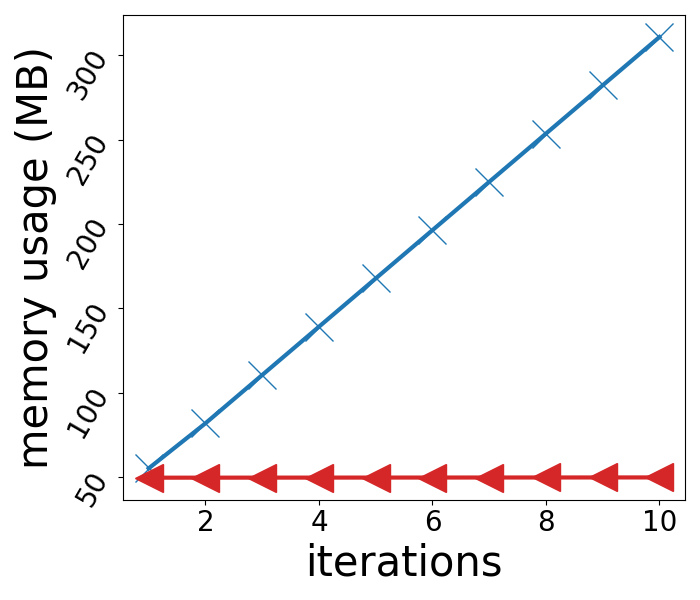}
		\\ (c) 10 iterations
		& (d) problem size 500
	\end{tabular}
	\caption{Time and memory comparison for optimal transport.
		Our block-inverse implicit differentiation is much faster than full-inverse version and uses less memory than autograd.}
	\label{fig:ot_runtime_memory}
\end{figure}

\renewcommand{\floatpagefraction}{0.9}
\begin{figure*}
\begin{pycode}
def backward(ctx, dJdP):
	M, r, c, P = ctx.saved_tensors
	b, m = M.shape[:2]

	# initialize backward gradients (-v^T H^{-1} B with v = dJdP and B = I)
	dJdM = -1.0 * ctx.gamma * P * dJdP

	# compute [vHAt1, vHAt2] = v^T H^{-1} A^T as two blocks
	vHAt1, vHAt2 = dJdM[:, 1:m].sum(dim=2), dJdM.sum(dim=1)

	# compute [v1,v2] = -v^T H^{-1} A^T (A H^{-1] A^T)^{-1} by block inverse
	PdivC = P[:, 1:m] / c.view(b, 1, -1)
	block_11 = cholesky(diag_embed(r[:, 1:m]) - einsum("bij,bkj->bik", P[:, 1:m], PdivC))
	block_12 = cholesky_solve(PdivC, block_11)
	block_22 = diag_embed(1/c) + einsum("bji,bjk->bik", block_12, PdivC)

	v1 = cholesky_solve(vHAt1.view(b,m-1,1), block_11).view(b,m-1) - \
		einsum("bi,bji->bj", vHAt2, block_12)
	v2 = einsum("bi,bij->bj", vHAt2, block_22) - einsum("bi,bij->bj", vHAt1, block_12)

	# compute v^T H^{-1} A^T (A H^{-1] A^T)^{-1} A H^{-1} B - v^T H^{-1} B
	dJdM[:, 1:m] -= v1.view(b, m-1, 1) * P[:, 1:m]
	dJdM -= v2.view(b, 1, -1) * P
				
	# compute -v^T H^{-1} A^T (A H^{-1] A^T)^{-1} C for r and c
	dJdr = -1.0 / ctx.gamma * cat((zeros(b, 1), v1), dim=1)
	dJdc = -1.0 / ctx.gamma * v2

	return dJdM, dJdr, dJdc	
\end{pycode}
\caption{Implementation of the backward pass for optimal transport. Assumes that the inputs $M$, $r$ and $c$, and output $P$ are cached by the forward pass. Input $M$ and output $P$ consist of $b$ batches of $m$-by-$n$ matrices. Full source code available at \url{http://deepdeclarativenetworks.com}.}
\label{code:optimal_transport}
\end{figure*}

A PyTorch implementation for the gradient is shown in \figref{code:optimal_transport}. Here we evaluate the expression for the gradient from left-to-right and replace explicit multiplication by $A$ with corresponding summations of terms in the multiplicand (see Line 9). Rather than flattening $P$ we keep it in tensor form. Line 6 initializes the calculation of $\dd[M] J$ with $-v\transpose H^{-1} B$. This can be seen as an approximation to the gradient with constraints ignored and is close to the true gradient when only a small number of Sinkhorn iterations is needed in the forward pass.

Profiling this approximation is included in our experiments, where we also compare block inverse of $AH^{-1}A\transpose$ versus the full inverse (see \figref{fig:ot_runtime_memory}). Important to observe is that unrolling Sinkhorn (autograd) and the implicit differentiation approach with block inverse have approximately the same running time (\figref{fig:ot_runtime_memory}(a) and (b)) whereas the latter is much more memory efficient, improving over unrolling Sinkhorn beyond four iterations for problems of size 500-by-500 (\figref{fig:ot_runtime_memory}(d)).

We can similarly back propagate through $r$ and $c$ ($c$ omitted for brevity). Here we note that
\begin{align}
	\frac{\partial f}{\partial r_i} &= -\frac{1}{\gamma r_i} \sum_{j=1}^{n} P_{ij} = -\frac{1}{\gamma}
	\, \implies \,
	\frac{\partial^2 f}{\partial r_i \partial P_{kl}} = 0.
\end{align}
As such $B = 0$ and the expression in \eqnref{eqn:dydx} reduces to
\begin{align}
	-H^{-1} A\transpose \left(A H^{-1} A\transpose\right)^{-1} C,
\end{align}
where, by inspection of the constraint function in \eqnref{eqn:op_constraints},
\begin{align}
	C &=
	\begin{bmatrix}
		\frac{\partial h_p}{\partial r_i} \mid p = 1, \ldots, m+n-1,\, i = 1, \ldots, m
	\end{bmatrix} \\
	&=
	- \begin{bmatrix}
		I_{{m-1} \times m} \\
		0_{n \times m}
	\end{bmatrix} \in \reals^{(m + n - 1) \times m}.
\end{align}

The calculation of $v\transpose H^{-1}A\transpose(AH^{-1}A\transpose)^{-1}$ can be reused for the gradients associated with $M$, $r$ and $c$ as done in Lines~22 and 23 of the code. Note that taking a step in the (negative) gradient direction may destroy normalization of $r$ (or $c$) required by the optimal transport problem. One way to ensure normalization is preserved is to define $r$ in terms of another positive vector $\tilde{r}$ as $r = \tilde{r} / \ones\transpose \tilde{r}$. The backward going gradient would then need to be post-multiplied by
%
%
$\dd r(\tilde{r}) = (I_{n \times n} - r \ones\transpose) / \ones\transpose \tilde{r}$,
omitted in \figref{code:optimal_transport} for simplicity of exposition.


\section{Discussion}
\label{sec:discussion}

In this paper we studied two examples of deep declarative nodes and showed how to implement an efficient backward pass by exploiting problem structure. This results in better utilization of memory and compute than can be achieved from automatic differentiation (autodiff) or unrolling the forward pass optimization loop. However, for other problems unrolling or autodiff may be satisfactory for a given task despite being computationally more expensive. We now summarize several key practical implementation considerations for developing new deep declarative nodes if compute is an issue, using our case studies as a guide.

It is judicious to first implement and experiment with the declarative node using a generic automatic differentiation approach. Several open-source tools make this easy~\cite{Gould:PAMI2021, Agrawal:NIPS2019, Blondel:TR2021}. Moreover, having such an implementation allows for rapid testing of new ideas and will facilitate debugging of future specialized code in addition to the use of numerical gradient checking (\eg \texttt{\small autograd.gradcheck}).

Next, inspect the required derivatives for structure and use this to simplify the computation. For example, efficient algorithms exist for inverting certain Hessian matrices (diagonal or block, triangular, etc.), and multiplication by 0-1 matrices can be replaced with summations. Importantly, when the objective of the problem decomposes elementwise over the optimization variables, such as in optimal transport, then the Hessian matrix will be diagonal. Related to this is thinking about the order of operations in \eqnref{eqn:dydx}, which can dramatically affect the memory required for storing intermediate results. The vector-Jacobian product used for computing the loss in the backward pass is a good example of this, as is the left-to-right evaluation of the outer products required for robust vector pooling, which is common when norms appear in objective or constraint functions.

Other standard considerations include saving calculations in the forward pass (if tractable to do so); disabling autodiff in the forward pass, which avoids unnecessary construction of the computation graph; performing inline operations to reuse memory buffers; and batch operations for better parallelism. Numerical stability can also be an issue, especially when the (locally) optimal solution is not isolated or the Hessian is almost singular. Here, linear system solvers (\eg Cholesky) should be used instead of inverting matrices and trust-region approaches (or regularization of the Hessian) can be used to improve stability~\cite{Toso:NIPS2019, Gould:PAMI2021}.

Finally, reparametrizing the problem can give different computational trade-offs, \eg removing constraints to make a problem unconstrained or adding variables (and associated constraints) so that the Hessian is structured. Alternatively, taking a hybrid approach where structure is exploited for some terms and autodiff used for the rest. This is particularly attractive when the optimality conditions can be written as the composition of many functions (as was done for example in \citename{Campbell:ECCV2020}). Moreover, it presents an exciting future research direction to see whether some of these techniques can be applied automatically.


{
  \bibliography{long,papers}

\begin{thebibliography}{19}
\providecommand{\natexlab}[1]{#1}

\bibitem[{Agrawal et~al.(2019)Agrawal, Amos, Barratt, Boyd, Diamond, and
  Kolter}]{Agrawal:NIPS2019}
Agrawal, A.; Amos, B.; Barratt, S.; Boyd, S.~P.; Diamond, S.; and Kolter, Z.
  2019.
\newblock Differentiable Convex Optimization Layers.
\newblock In \emph{Advances in Neural Information Processing Systems
  ({NeurIPS})}.

\bibitem[{Amos and Kolter(2017)}]{Amos:ICML2017}
Amos, B.; and Kolter, Z. 2017.
\newblock {O}pt{N}et: Differentiable Optimization as a Layer in Neural
  Networks.
\newblock In \emph{Proc.~of the International Conference on Machine Learning
  ({ICML})}.

\bibitem[{Asano, Rupprecht, and Vedaldi(2020)}]{Asano:ICLR2021}
Asano, Y.~M.; Rupprecht, C.; and Vedaldi, A. 2020.
\newblock Self-labelling via simultaneous clustering and representation
  learning.
\newblock In \emph{Proc.~of the International Conference on Learning
  Representations ({ICLR})}.

\bibitem[{Blondel et~al.(2021)Blondel, Berthet, Cuturi, Frostig, Hoyer,
  Llinares-Lopez, Pedregosa, and Vert}]{Blondel:TR2021}
Blondel, M.; Berthet, Q.; Cuturi, M.; Frostig, R.; Hoyer, S.; Llinares-Lopez,
  F.; Pedregosa, F.; and Vert, J.-P. 2021.
\newblock Efficient and Modular Implicit Differentiation.
\newblock Technical report, Google (arXiv:2105.15183).

\bibitem[{Boyd and Vandenberghe(2004)}]{Boyd:2004}
Boyd, S.~P.; and Vandenberghe, L. 2004.
\newblock \emph{Convex Optimization}.
\newblock Cambridge.

\bibitem[{Campbell, Liu, and Gould(2020)}]{Campbell:ECCV2020}
Campbell, D.; Liu, L.; and Gould, S. 2020.
\newblock Solving the Blind Perspective-n-Point Problem End-To-End with Robust
  Differentiable Geometric Optimization.
\newblock In \emph{Proc.~of the European Conference on Computer Vision
  ({ECCV})}.

\bibitem[{Chen et~al.(2020)Chen, Parra, Cao, Li, and Chin}]{Chen:CVPR2020}
Chen, B.; Parra, A.; Cao, J.; Li, N.; and Chin, T.-J. 2020.
\newblock End-to-End Learnable Geometric Vision by Backpropagating {PnP}
  Optimization.
\newblock In \emph{Proc.~of the {IEEE} Conference on Computer Vision and
  Pattern Recognition ({CVPR})}.

\bibitem[{Cuturi(2013)}]{Cuturi:NIPS2013}
Cuturi, M. 2013.
\newblock Sinkhorn Distances: Lightspeed Computation of Optimal Transport.
\newblock In \emph{Advances in Neural Information Processing Systems
  ({NeurIPS})}.

\bibitem[{Diamond and Boyd(2016)}]{cvxpy}
Diamond, S.; and Boyd, S. 2016.
\newblock {CVXPY}: {A} {P}ython-embedded modeling language for convex
  optimization.
\newblock \emph{Journal of Machine Learning Research}, 17(83): 1--5.

\bibitem[{Fernando et~al.(2016)Fernando, Anderson, Hutter, and
  Gould}]{Fernando:CVPR2016}
Fernando, B.; Anderson, P.; Hutter, M.; and Gould, S. 2016.
\newblock Discriminative Hierarchical Rank Pooling for Activity Recognition.
\newblock In \emph{Proc.~of the {IEEE} Conference on Computer Vision and
  Pattern Recognition ({CVPR})}.

\bibitem[{Fernando and Gould(2016)}]{Fernando:ICML2016}
Fernando, B.; and Gould, S. 2016.
\newblock Learning End-to-end Video Classification with Rank-Pooling.
\newblock In \emph{Proc.~of the International Conference on Machine Learning
  ({ICML})}.

\bibitem[{Gould et~al.(2016)Gould, Fernando, Cherian, Anderson, {Santa Cruz},
  and Guo}]{Gould:TR2016}
Gould, S.; Fernando, B.; Cherian, A.; Anderson, P.; {Santa Cruz}, R.; and Guo,
  E. 2016.
\newblock On Differentiating Parameterized Argmin and Argmax Problems with
  Application to Bi-level Optimization.
\newblock Technical report, Australian National University (arXiv:1607.05447).

\bibitem[{Gould, Hartley, and Campbell(2021)}]{Gould:PAMI2021}
Gould, S.; Hartley, R.; and Campbell, D. 2021.
\newblock Deep Declarative Networks.
\newblock \emph{{IEEE} Trans.~on Pattern Analysis and Machine Intelligence
  ({PAMI})}.

\bibitem[{Horn and Johnson(1991)}]{Horn:1991}
Horn, R.~A.; and Johnson, C.~R. 1991.
\newblock \emph{Topics in Matrix Analysis}.
\newblock Cambridge University Press.

\bibitem[{Lee et~al.(2019)Lee, Maji, Ravichandran, and Soatto}]{Lee:CVPR2019}
Lee, K.; Maji, S.; Ravichandran, A.; and Soatto, S. 2019.
\newblock Meta-Learning with Differentiable Convex Optimization.
\newblock In \emph{Proc.~of the {IEEE} Conference on Computer Vision and
  Pattern Recognition ({CVPR})}.

\bibitem[{Luise et~al.(2018)Luise, Rudi, Pontil, and
  Ciliberto}]{Luise:NIPS2018}
Luise, G.; Rudi, A.; Pontil, M.; and Ciliberto, C. 2018.
\newblock Differential Properties of Sinkhorn Approximation for Learning with
  Wasserstein Distance.
\newblock In \emph{Advances in Neural Information Processing Systems
  ({NeurIPS})}, volume~31.

\bibitem[{Paszke et~al.(2017)Paszke, Gross, Chintala, Chanan, Yang, DeVito,
  Lin, Desmaison, Antiga, and Lerer}]{PyTorch}
Paszke, A.; Gross, S.; Chintala, S.; Chanan, G.; Yang, E.; DeVito, Z.; Lin, Z.;
  Desmaison, A.; Antiga, L.; and Lerer, A. 2017.
\newblock Automatic Differentiation in {PyTorch}.
\newblock In \emph{NeurIPS Autodiff Workshop}.

\bibitem[{Toso, Campbell, and Russell(2019)}]{Toso:NIPS2019}
Toso, M.; Campbell, N.; and Russell, C. 2019.
\newblock Fixing Implicit Derivatives: Trust-Region Based Learning of
  Continuous Energy Functions.
\newblock In \emph{Advances in Neural Information Processing Systems
  ({NeurIPS})}.

\bibitem[{Wang et~al.(2019)Wang, Donti, Wilder, and Kolter}]{Wang:ICML2019}
Wang, P.-W.; Donti, P.~L.; Wilder, B.; and Kolter, Z. 2019.
\newblock {SATN}et: Bridging deep learning and logical reasoning using a
  differentiable satisfiability solver.
\newblock In \emph{Proc.~of the International Conference on Machine Learning
  ({ICML})}.

\end{thebibliography}
}

\end{document}